\documentclass[10pt,twocolumn,letterpaper]{article}

\usepackage{iccv}
\usepackage{times}
\usepackage{epsfig}
\usepackage{epstopdf}
\usepackage{graphicx}
\usepackage{amsmath}
\usepackage{amssymb}
\usepackage{subfigure}
\usepackage{bm}
\usepackage{algorithmic}
\usepackage[ruled]{algorithm2e}

\usepackage[pagebackref=true,breaklinks=true,letterpaper=true,colorlinks,bookmarks=false]{hyperref}

\begin{document}

\title{Graph Contrastive Clustering}

\author{

Huasong Zhong$^{1}$\thanks{\scriptsize{Equal contribution and this work was done when Huasong Zhong worked full time at DAMO Academy, Alibaba Group}}, Jianlong Wu$^{2*}$, Chong Chen$^{1,3}$\thanks{Corresponding author.}, Jianqiang Huang$^{1}$ \\ 
  Minghua Deng$^{3}$, Liqiang Nie$^{2}$, Zhouchen Lin$^{3}$, Xian-Sheng~Hua$^{1}$  \\

$^1$DAMO Academy, Alibaba Group  \hspace{1cm} $^2$Shandong University \hspace{1cm} $^3$Peking University  \\

{\tt\small \{huasong.zhs, cheung.cc, jianqiang.hjq\}@alibaba-inc.com }, {\tt\small jlwu1992@sdu.edu.cn} \\
{\tt\small  \{huaxiansheng, nieliqiang\}@gmail.com }, {\tt\small \{dengmh, zlin\}@pku.edu.cn } 

}
\date{}
\maketitle

\begin{abstract}
Recently, some contrastive learning methods have been proposed to simultaneously learn representations and clustering assignments, achieving significant improvements. However, these methods do not take
the category information and clustering objective into consideration, thus the learned representations are not optimal for clustering and the performance might be limited. Towards this issue, we first propose a novel graph contrastive learning framework, 
and then apply it to the clustering task, resulting in 
the Graph Constrastive Clustering~(GCC) method. Different from basic contrastive clustering that only assumes an image and its augmentation should share similar representation and clustering assignments, we lift the instance-level consistency to the cluster-level consistency with the assumption that samples in one cluster and their augmentations should all be similar. Specifically, on the one hand, we propose the graph Laplacian based contrastive loss to learn more discriminative and clustering-friendly features. On the other hand, we propose a novel graph-based contrastive learning strategy to learn more compact clustering assignments. Both of them incorporate the latent category information to reduce the intra-cluster variance as well as increase the inter-cluster variance. Experiments on six commonly used datasets demonstrate the superiority of our proposed approach over the state-of-the-art methods.
\end{abstract}

\section{Introduction}
Based on a large number of annotated training samples,
deep learning achieves significant success in the past decade~\cite{resnet}.
However, it is very expensive and time-consuming to manually label a large training dataset.
It is also impractical to collect a labeled dataset for each domain or task.
In this case, clustering attracts more attention recently,
which aims to divide the samples into separate clusters without knowing the label information.
\par
\begin{figure}[t]
	\centering
	\includegraphics[width=\linewidth]{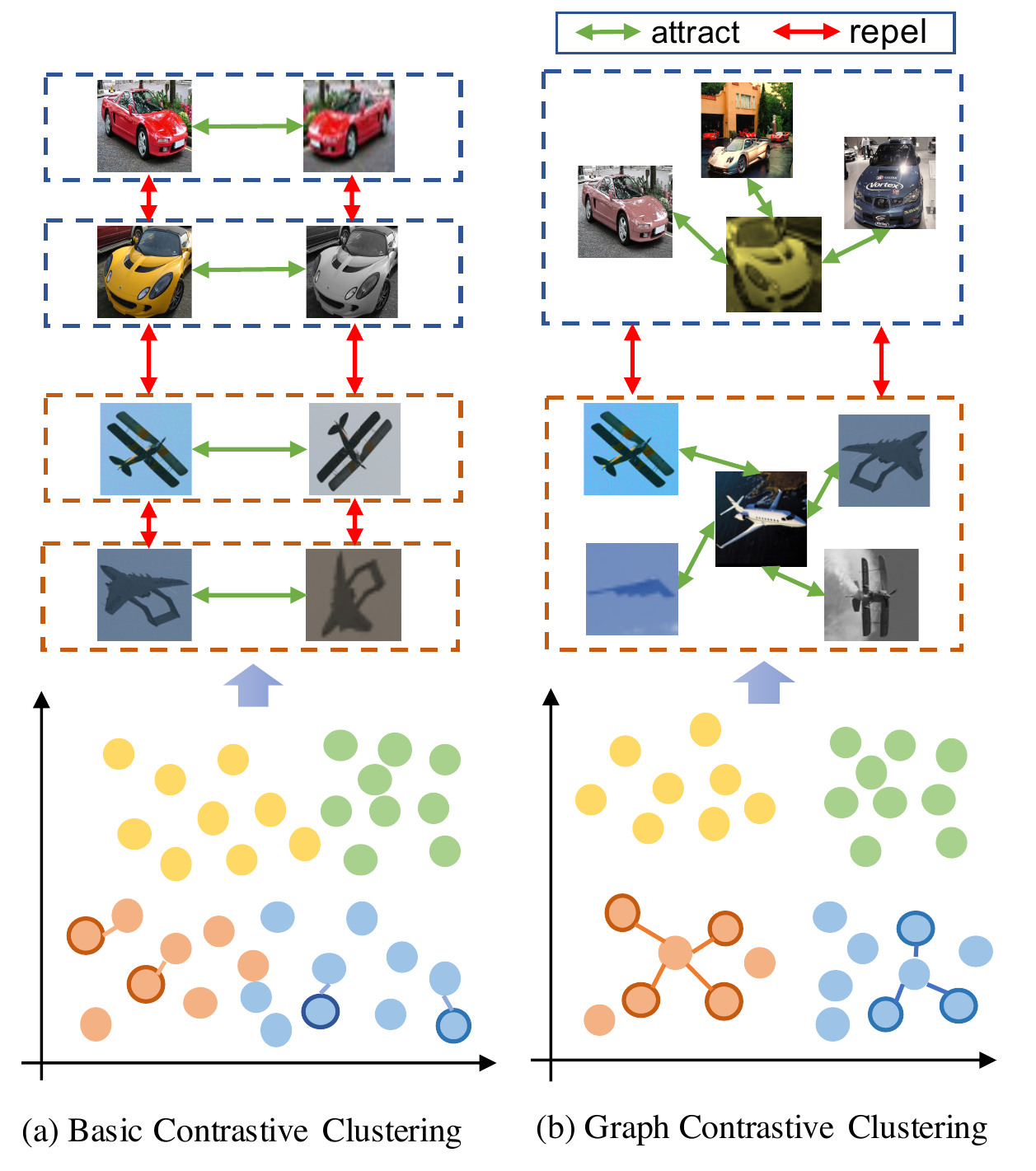}
	\caption{Motivation of the proposed GCC. (a) Existing contrastive learning based clustering methods mainly focus on instance-level consistency, which maximizes the correlation between self-augmented samples and treats all other samples as negative samples. (b) GCC incorporates the category information to perform the contrastive learning at both the instance and the cluster levels, which can better minimize the intra-cluster variance and maximize the inter-cluster variance.}
	\label{fig:pre}
\end{figure}
Clustering~\cite{dac,dccm,pica,iic} is a very challenging task since samples in the same class have various appearances and supervision signals are lacked to train the model.
Classic clustering methods~~\cite{zelnik2005self, gowda1978agglomerative, cai2009locality}, such as spectral clustering~\cite{ng2002spectral} and subspace clustering~\cite{liu2013robust, elhamifar2009sparse}, suffer from two obvious limitations, including indiscriminative feature representation and sub-optimal solution for clustering caused by the separation of feature extraction and clustering. Some recent deep learning based methods can well handle the above issues.
For example, auto-encoder related methods~\cite{vincent2010stacked,kingma2013auto} minimize the reconstruction error and assign various regularization terms in the latent feature space, such as the KL-divergence~\cite{xie2016unsupervised}.
Deep adaptive clustering~(DAC)~\cite{dac} maximizes the similarity between self-augmented samples to adaptively train the neural network. Deep comprehensive correlation mining~(DCCM)~\cite{dccm} thoroughly investigates various kinds of correlation among samples and features. These approaches achieve good clustering performance, but their upper bound accuracy is limited since the learned features are not discriminative enough. 
\par
Recently, contrastive learning~\cite{simCLR} has received much attention in unsupervised feature learning, which emphasizes the importance of data augmentation and maximizes the agreement between two augmented samples. 
Because of its success, a few approaches~\cite{pica,drc,cc,dccm} are proposed to jointly optimize the contrastive learning and clustering.
For instance, partition confidence maximisation~(PICA)~\cite{pica} learns the most semantically plausible clustering solution by maximizing partition confidence, which corresponds to the cluster-wise contrastive learning.
Instead of only using the cluster contrast in PICA, 
deep robust clustering~(DRC)~\cite{drc} adopts the conventional contrastive learning in feature and cluster space simultaneously.
These methods significantly improve the clustering performance, but they still face another obvious issue: both of them still follow the basic framework of contrastive learning and only assume that a sample and its augmentations should be similar in the feature space, which does not incorporate the latent category information into clustering.
\par
In view of the above limitations, we propose the graph contrastive framework and apply it to the clustering task, resulting in the Graph Contrastive Clustering~(GCC) method.
As shown in Figure~\ref{fig:pre}, we assume that samples in one cluster and their augmentations should share similar feature representations and clustering assignments, which lifts the commonly-used instance-level consistency in PICA and DRC to the cluster-level consistency. By incorporating the latent category/cluster information, GCC can help to learn more discriminative features and better clustering assignments, which is more suitable for the clustering task. Specifically, we first construct a similarity graph based on the current features, then we apply it to both representation learning and clustering learning. For representation learning, the graph Laplacian based contrastive loss is proposed to learn more clustering-friendly features. For clustering learning, a novel graph-based contrastive learning strategy is proposed to learn more compact clustering assignments. Both of them can help to decrease the intra-class variance and increase inter-class variance.
Experimental results on six challenging datasets validate the effectiveness of the proposed method. We also perform extensive ablation analysis to demonstrate the superiority of graph contrastive. 
\par
The contributions of this paper can be summarized as follows:
\begin{enumerate}
	\item By incorporating the latent category information, we propose a novel graph contrastive framework, which assumes that samples in one cluster and their augmentations should share similar representations and clustering assignments. This framework lifts the tradition instance-level consistency to cluster-level consistency, thus can better reduce the intra-class variance as well as increase the inter-class variance.
	\item We apply the proposed graph contrastive framework to the clustering task, and come up with the graph contrastive clustering method~(GCC), which consists of two graph contrastive modules. For representation graph contrastive module, a graph Laplacian based contrastive loss is proposed to learn more discriminative and clustering-friendly features. For assignment graph contrastive module, a novel graph-based contrastive learning strategy is proposed to learn more compact clustering assignments.
	\item We conduct extensive experiments on image clustering and our proposed method achieves significant improvement on various datasets. We also conduct an extensive ablation study to validate the effectiveness of each proposed module.
\end{enumerate}

\section{Related work}

\subsection{Deep Clustering}
According the difference in self-supervised signal,
deep clustering methods can be mainly divided into two categories, including the reconstruction based methods~\cite{xie2016unsupervised,peng2016deep, depict,guo2017improved,yang2017towards} and the self-augmentation based methods~\cite{dac, dccm, iic, haeusser2018associative, pica, scan, drc}.

The former adopts the auto-encoder~\cite{vincent2010stacked} framework and imposes different regularization terms on the latent feature learning.
For example, DEC~\cite{xie2016unsupervised} and IDEC~\cite{guo2017improved} minimize the KL-divergence for features in the latent subspace. 
Peng et al.~\cite{peng2016deep} incorporate the sparsity prior.
Yang et al.~\cite{yang2017towards} combine it with K-means.
DEPICT~\cite{depict} proposes the relative entropy minimization based on convolutional auto-encoder.
The latter focuses on exploiting the consistent information between original images and their transformed images to train the network. 
DAC~\cite{dac} adopts a binary pairwise classification framework for image clustering to make the feature learning in a “supervised” manner. 
DCCM~\cite{dccm} comprehensively utilizes various kinds of correlations among representations.
IIC~\cite{iic} maximizes the mutual information of positive pairs to make them keep a similar assignment probability.   
PICA~\cite{pica} learns the most semantically plausible clustering solution by maximizing partition confidence.  
DRC~\cite{drc}  tries to learn invariant features and clusters by introducing contrastive learning to optimize the consistency between image and its augmentation simultaneously. 
SCAN~\cite{scan} utilizes a three-stage method to improve the clustering.
These approaches achieve good results, but they ignore the connections between cluster assignment learning and representation learning. 
As a contrast, our method considers their connections, and simultaneously learn both feature representation and cluster assignment.

\subsection{Contrastive Learning}
Recently, constrastive learning achieves significant progress, and it can learn discriminative feature representation without any manual annotations.
For example, Wu et al.~\cite{wu2018unsupervised} introduces a memory bank to store the embedding of instance representation. 
Zhuang et al.~\cite{zhuang2019local} extends the above memory bank by learning an embedding function to maximize a metric of local aggregation, causing similar data instances to move together in the embedding space. MoCo~\cite{MoCo} views contrastive learning as dictionary loop-up and built
a dynamic dictionary with a queue and a moving-averaged encoder. 
MoCo v2~\cite{chen2020improved} makes simple modifications to MoCo by using an MLP projection head and more data augmentations. 
simCLR~\cite{simCLR} simplifies recently proposed contrastive self-supervised learning algorithms without requiring specialized architectures or a memory bank. simCLR v2~\cite{chen2020big} finds that bigger self-supervised models are more label efficient, performing significantly better when fine-tuned on only a few labeled examples, even though they have more capacity to potentially overfit.
Tian et al.~\cite{tian2019contrastive, tian2020ICLR} extends the constrastive learning to the multi-view case and representation distillation.
Although these methods can learn good feature representations, how to apply them to the clustering task to improve the performance still remains challenging.

\section{Graph Contrastive Clustering}

\begin{figure*}
	\centering
	\includegraphics[width=\linewidth]{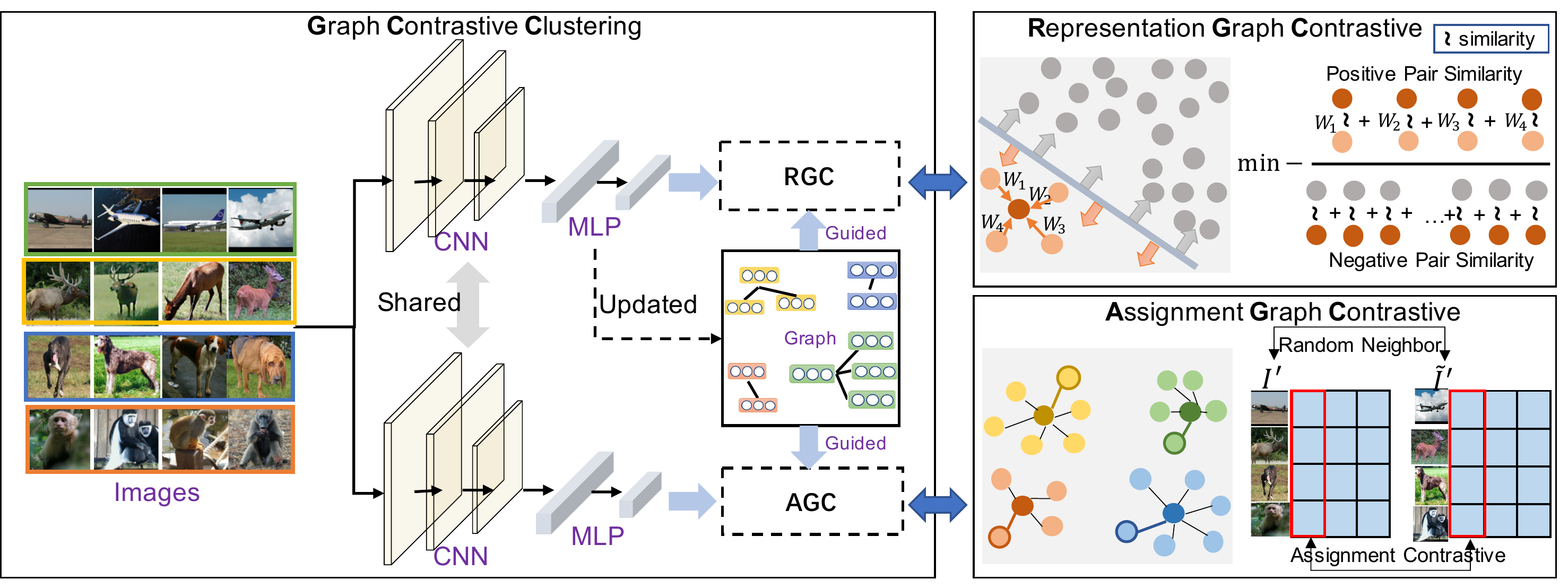}
	\caption{
	Framework of the proposed Graph Contrastive Clustering. GCC has two heads with shared CNN parameters. The first head is a representation graph contrastive (RGC) module, which helps to learn clustering-friendly features. The second head is an assignment graph contrastive (AGC) module, which leads to a more compact cluster assignment.
	} 
	\label{fig:main_overview} 
\end{figure*}

\subsection{Problem Formulation}
Given a set of $N$ unlabelled images $\mathbf{I} = \{I_{1}, ..., I_{N}\}$ from $K$ different categories, deep clustering aims to separate these images into $K$ different clusters by convolutional neural network (CNN) models such that the images with the same semantic labels can be grouped into the same cluster. Here we aim to learn a deep CNN network based mapping function $\Phi$ with parameter $\theta$, such that each image $I_{i}$ can be mapped to $(z_{i}, p_{i})$, where $z_{i}$ is the $d$-dimensional representation feature with regularization $\|z_{i}\|_{2} = 1$ and $p_{i}$ is the $K$-dimension assignment probability which satisfies $\sum_{j=1}^{K}p_{ij} = 1$. Then the cluster assignment for the $i$-th sample~($i = 1,...,N$) can be predicted by the following maximum likelihood:
$$
\ell_{i} = \arg\max_{j}(p_{ij}), 1\leq j\leq K.
$$

\subsection{Graph Contrastive (GC)}
Let $G=(V, E)$ be an undirected  graph with vertex set $V=\{v_{1},\cdots, v_{N}\}$. The edge set $E$ can be represented by the adjacency matrix $A$ such that:
\begin{equation}
    A_{ij} = \begin{cases}
    1, & \text{if} \ (v_{i}, v_{j}) \in E; \\
    0, & \text{otherwise}.
    \end{cases}
\end{equation}
Let $d_{i}$ be the degree of $v_{i}$, if we define $D = 
\begin{bmatrix}
d_{1} & \cdots & 0 \\
\vdots & \ddots & \vdots \\
0 & \cdots & d_{n}
\end{bmatrix}
$, then the normalized symmetric Graph Laplacian of $G$ can be defined as:
\begin{equation}
\label{equ:laplacian}
L = I - D^{-\frac{1}{2}}AD^{-\frac{1}{2}}.
\end{equation}
It is easy to check that $L_{ij} = -\frac{A_{ij}}{\sqrt{d_{i}d_{j}}}, i\neq j$.

Given $N$ representation features $\mathbf{x} = \{x_{1}, ..., x_{N}\}$ with unit $\ell_{2}$ norm, the intuition of GC is that $x_{i}$ should be close to $x_{j}$ if $A_{ij} > 0$ while $x_{i}$ should be far away from $x_{j}$ if $A_{ij} = 0$. Assume that the graph can be partitioned into several communities, the intuition of GC tells us that the similarities of feature representations  in the same community should be larger than that between communities. Approximately, we can define
\begin{equation}
    \mathcal{S}_{intra} = \sum_{L_{ij} < 0}-L_{ij}S(x_{i}, x_{j})
\end{equation}
as the total intra-community similarity and 
\begin{equation}
    \mathcal{S}_{inter} = \sum_{L_{ij} = 0}S(x_{i}, x_{j})
\end{equation}
as the total inter-community similarity, where $S(x_{i}, x_{j})$ is the similarity between $x_{i}$ and $x_{j}$. Then we can mathematically define the loss of GC as:
\begin{equation}
\label{equ:gcloss}
    \mathcal{L}_{GC} = -\frac{1}{N}\sum_{i=1}^{N}\log\left(\frac{\sum_{L_{ij}<0}-L_{ij}S(x_{i},x_{j})}{\sum_{L_{ij}=0}S(x_{i},x_{j})}\right).
\end{equation}
Minimizing $\mathcal{L}_{GC}$ can simultaneously increase total intra-community similarity and decrease total inter-community similarity, which can improve the separableness and lead to the result that learned feature representations are consistent with the graph structure.

\subsection{Framework of GCC}
We introduce a novel end-to-end deep clustering framework by applying GC to both representation learning and assignment learning. As shown in Figure \ref{fig:main_overview}, 
there are two heads with shared CNN parameters in our GCC model. The upper head is a representation graph contrastive (RGC) module, 
which learns clustering-friendly features based on representation graph contrastive learning. 
The bottom head is an assignment graph contrastive (AGC) module, which achieves the final cluster assignment with cluster-level graph contrastive learning. 
With these two modules, GCC can simultaneously learn more discriminative features and clusters to improve clustering.
We will present the details of GCC below.

\subsubsection{Graph Construction}
Since the deep learning model usually fluctuates during training, the representation features of an epoch may have large biases. We take advantage of moving average to reduce this kind of bias before graph construction. To be specific, assume that $\Phi_{\theta}^{(t)}$ is the model and $Z^{(t)} = (z_{1}^{(t)},\cdots,z_{N}^{(t)}) = (\Phi_{\theta}^{(t)}(I_{1}),\cdots, \Phi_{\theta}^{(t)}(I_{N}))$ are the representation features of $t$-th epoch, the moving average of representation features can be defined as:
$$\bar{z}_{i}^{(t)} = \frac{(1-\alpha)\bar{z}_{i}^{(t-1)} + \alpha z_{i}^{(t)}}{\|(1-\alpha)\bar{z}_{i}^{(t-1)} + \alpha z_{i}^{(t)}\|_{2}}, i = 1,\cdots, N, $$
where $\alpha$ is a parameter to trade-off current and past effects and $\bar{z}_{i}^{(0)} = z_{i}^{(0)}$. Then we can construct the KNN graph by
\begin{equation}
\label{equ:knn}
A_{ij}^{(t)}=
\begin{cases}
1, & \text{if} \ \bar{z}_{j}^{(t)}\in \mathcal{N}^{k}(\bar{z}_{i}^{(t)}) \text{ or } \  \bar{z}_{i}^{(t)}\in \mathcal{N}^{k}(\bar{z}_{j}^{(t)});\\
0, & \text{otherwise}
\end{cases}
\end{equation}
for $i,j=1,\cdots,N$. After that, the Graph Laplacian $L^{(t)}$ can be obtained by Eq.~(\ref{equ:laplacian}). 

\subsubsection{Similarity Function}
To compute the similarity between two samples, we adopt the Gaussian kernel function which is commonly used in spectral clustering.
The similarity in GC loss Eq.~(\ref{equ:gcloss}) can be defined as:
$$
S(x_{i},x_{j}) = e^{-\|x_{i}-x_{j}\|_{2}^{2}/\tau},
$$
where $\tau$ is a parameter that represents variance or temperature. Since $\|x_{i}-x_{j}\|_{2}^{2} = \|x_{i}\|_{2}^{2} + \|x_{j}\|_{2}^{2} - 2x_{i}\cdot x_{j} = 2 - 2x_{i}\cdot x_{j}$, we use the following similarity function as a substitution:
\begin{equation}
S(x_{i},x_{j}) = e^{x_{i}\cdot x_{j}/\tau}.
\end{equation}

\subsubsection{Representation Graph Contrastive}

Assume $\mathbf{I}^{'} = \{I_{1}^{'}, ..., I_{N}^{'}\}$ is a random transformation of original images, and their corresponding features are $\mathbf{z}^{'} = (z_{1}^{'},\cdots,z_{N}^{'})$. According to graph contrastive mentioned before, $z_{i}^{'}$ and $z_{j}^{'}$ should be similar if they are linked while be far away if they are disconnected. Let $\mathbf{x} = \mathbf{z}^{'}$ in Equation (\ref{equ:gcloss}), we can get the loss of RGC learning as:

\begin{equation}
\label{equ:rgcloss}
    \mathcal{L}_{RGC}^{(t)} = -\frac{1}{N}\sum_{i=1}^{N}\log\left(\frac{\sum_{L_{ij}^{(t)}<0}-L_{ij}^{(t)}e^{z_{i}^{'}\cdot z_{j}^{'}/\tau}}{\sum_{L_{ij}=0}e^{z_{i}^{'}\cdot z_{j}^{'}/\tau}}\right).
\end{equation}

\subsubsection{Assignment Graph Contrastive}

For traditional contrastive learning based clustering, images and their augmentations should share similar cluster assignment distribution, \textit{e.g.}\ the index of images and their augmentations assigned to cluster $k$ should be consistent. It is reasonable but does not take advantage of clustering information. As the model gets better and better during training, images and their neighbors also should share similar cluster assignment distribution with high probability. Due to this motivation, we propose the assignment graph contrastive learning.

Assume that $\mathbf{I}^{'} = \{I_{1}^{'}, ..., I_{N}^{'}\}$ are the random augmentations of original images and $\tilde{\mathbf{I}}^{'} = \{\tilde{I}_{1}^{'}, ..., \tilde{I}_{N}^{'}\}$ satisfies that $\tilde{I}_{j}^{'}$ is a transformation of a random neighbor of $I_{i}$ according to graph $A^{(t)}$, the assignment probability matrix for $\mathbf{I}^{'}$ and $\tilde{\mathbf{I}}^{'}$ can be defined as
$$
\begin{gathered}
\mathbf{p}^{'}=
\begin{bmatrix} 
p_{1}^{'} \\ 
... \\
p_{N}^{'}
\end{bmatrix}_{N\times K}
\text{ and }
\tilde{\mathbf{p}}^{'}=
\begin{bmatrix} 
p_{\text{RN}(I_{1})}^{'} \\ 
... \\
p_{\text{RN}(I_N)}^{'}
\end{bmatrix}_{N\times K}
\end{gathered},
$$
where $\text{RN}(I_{i})$ denotes a random neighbor of image $I_{i}$. We can reformulate them by the following column vector forms:
$$
\begin{gathered}
\mathbf{q}^{'}=
\begin{bmatrix} 
q^{'}_{1}, & ... &, q^{'}_{K}
\end{bmatrix}_{N\times K},
\end{gathered}
$$
$$
\begin{gathered}
\tilde{\mathbf{q}}^{'}=
\begin{bmatrix} 
\tilde{q}_{1}^{'}, & ... &, \tilde{q}_{K}^{'}
\end{bmatrix}_{N\times K},
\end{gathered}
$$
where $q_{i}^{'}$ and $\tilde{q}_{i}^{'}$ can tell us which pictures in $\mathbf{I}^{'}$ and $\tilde{\mathbf{I}}^{'}$ will be assigned to cluster $i$, respectively.  Then we can define the AGC learning loss as:
\begin{equation}
\label{equ:agcloss}
    \mathcal{L}_{AGC} = -\frac{1}{K}\sum_{i=1}^{K}\log\left(\frac{e^{q_{i}^{'}\cdot \tilde{q}_{i}^{'}/\tau}}{\sum_{j=1}^{K}e^{q_{i}^{'}\cdot \tilde{q}_{j}^{'}/\tau}}\right).
\end{equation}
\subsubsection{Cluster Regularization Loss}
In deep clustering, it is easy to fall into a local optimal solution that assign most samples into a minority of clusters. To avoid trivial solution, we also add a clustering regularization loss similar to PICA~\cite{pica} and SCAN~\cite{scan}:
\begin{equation}
\label{equ:crloss}
    \mathcal{L}_{CR} = \log(K) - H(\mathcal{Z}),
\end{equation}
where  $\mathcal{Z}_{i} \!=\! \frac{\sum_{j=1}^{N}q_{ij}}{\sum_{i=1}^{K}\sum_{j=1}^{N}q_{ij}}$, and
$
\begin{gathered}
\mathbf{q} = 
\begin{bmatrix} 
q_{1}, & \cdots &, q_{K}
\end{bmatrix}_{N\times K}
\end{gathered}
$
is the assign probability of $\mathbf{I}$.

Then the overall objective function of GCC can be formulated as:
\begin{equation}
\label{equ:obj}
    \mathcal{L} = \mathcal{L}_{RGC} + \lambda  \mathcal{L}_{AGC} + \eta  \mathcal{L}_{CR},
\end{equation}
where $\lambda$ and $\eta$ are weight parameters.

\subsection{Model Training}
The objective function in Eq.~\eqref{equ:obj} is differentiable and
end-to-end, enabling the conventional stochastic gradient descent algorithm for model training.  The training procedure is summarized in Algorithm~\ref{alg}.
\begin{algorithm}
\caption{Training algorithm for GCC \label{alg}}
\KwIn{Training images $\mathcal{I}=\{I_{1}, \dots, I_{N}\}$, training epochs $N_{ep}$, and number of clusters $K$.}
\KwOut{A deep clustering model with parameter $\Theta$.}
Initializing graph $A$ and parameter $\Theta$;\\
\For{each epoch}{
\textbf{Step 1: } Sampling a random mini-batch of images and their neighbors according to $A$;
\textbf{Step 2: } Generating augmentations for the sampled images and their neighbors;
\textbf{Step 3: } Computing representation graph contrastive loss according to Eq.~\eqref{equ:rgcloss};
\textbf{Step 4: } Computing assignment graph contrastive loss according to Eq.~\eqref{equ:agcloss};
\textbf{Step 5: } Computing cluster regularization loss according to Eq.~\eqref{equ:crloss};
\textbf{Step 6: } Update $\Theta$ with SGD by minimizing the overall loss according to Eq.~\eqref{equ:obj};\\
\textbf{Step 7:} Update $A$ according to Eq.~\eqref{equ:knn}.
}
\end{algorithm}

\section{Experiments}

\begin{table*}[!htbp]
    \centering
 	\renewcommand\arraystretch{1.05}
	\footnotesize
	\caption{Clustering performance of different methods on six challenging datasets.} 
 	\label{tab:clustering_res}
	\setlength{\tabcolsep}{2.0pt}
	\begin{tabular}{|c|ccc|ccc|ccc|ccc|ccc|ccc|}
		\hline
		Datasets                       & \multicolumn{3}{c|}{CIFAR-10}                       & \multicolumn{3}{c|}{CIFAR-100}                      & \multicolumn{3}{c|}{STL-10}       & \multicolumn{3}{c|}{ImageNet-10} & \multicolumn{3}{c|}{Imagenet-dog-15}                 & \multicolumn{3}{c|}{Tiny-ImageNet}                  \\ \hline
		Methods & NMI      & ACC    & ARI & NMI      & ACC     & ARI & NMI    & ACC        & ARI & NMI                   & ACC                   & ARI & NMI                   & ACC                   & ARI & NMI                   & ACC                   & ARI \\ \hline
		K-means & 0.087 & 0.229 & 0.049 & 0.084 & 0.130 & 0.028 & 0.125 & 0.192 & 0.061 & 0.119  & 0.241  & 0.057  & 0.055  & 0.105  & 0.020 & 0.065 & 0.025   &  0.005 \\ \hline
		SC      & 0.103 & 0.247 & 0.085 & 0.090 & 0.136 & 0.022 & 0.098 & 0.159 & 0.048 & 0.151  & 0.274  & 0.076  & 0.038  & 0.111  & 0.013 & 0.063  & 0.022   & 0.004  \\ \hline
		AC      & 0.105 & 0.228 & 0.065 & 0.098 & 0.138 & 0.034 & 0.239 & 0.332 & 0.140 & 0.138  & 0.242  & 0.067  & 0.037  & 0.139  & 0.021 & 0.069  & 0.027   & 0.005 \\ \hline
		NMF     & 0.081 & 0.190 & 0.034 & 0.079 & 0.118 & 0.026 & 0.096 & 0.180 & 0.046 & 0.132  & 0.230  & 0.065  & 0.044  & 0.118  & 0.016 & 0.072 & 0.029  & 0.005  \\ \hline
		AE      & 0.239 & 0.314 & 0.169 & 0.100 & 0.165 & 0.048 & 0.250 & 0.303 & 0.161 & 0.210  & 0.317  & 0.152  & 0.104  & 0.185  & 0.073 & 0.131 & 0.041  & 0.007 \\ \hline
		DAE     & 0.251 & 0.297 & 0.163 & 0.111 & 0.151 & 0.046 & 0.224 & 0.302 & 0.152 & 0.206  & 0.304  & 0.138  & 0.104  & 0.190  & 0.078 & 0.127 & 0.039   & 0.007  \\ \hline
		GAN     & 0.265 & 0.315 & 0.176 & 0.120 & 0.151 & 0.045 & 0.210 & 0.298 & 0.139 & 0.225  & 0.346  & 0.157  & 0.121  & 0.174  & 0.078 & 0.135 & 0.041   & 0.007  \\ \hline
		DeCNN   & 0.240 & 0.282 & 0.174 & 0.092 & 0.133 & 0.038 & 0.227 & 0.299 & 0.162 & 0.186  & 0.313  & 0.142  & 0.098  & 0.175  & 0.073 & 0.111 & 0.035   & 0.006  \\ \hline
		VAE     & 0.245 & 0.291 & 0.167 & 0.108 & 0.152 & 0.040 & 0.200 & 0.282 & 0.146 & 0.193  & 0.334  & 0.168  & 0.107  & 0.179  & 0.079 & 0.113 & 0.036  & 0.006 \\ \hline
		JULE    & 0.192 & 0.272 & 0.138 & 0.103 & 0.137 & 0.033 & 0.182 & 0.277 & 0.164 & 0.175  & 0.300  & 0.138  & 0.054  & 0.138  & 0.028 & 0.102 & 0.033  & 0.006 \\ \hline
		DEC     & 0.257 & 0.301 & 0.161 & 0.136 & 0.185 & 0.050 & 0.276 & 0.359 & 0.186 & 0.282  & 0.381  & 0.203  & 0.122  & 0.195  & 0.079 & 0.115  & 0.037   & 0.007  \\ \hline
		DAC     & 0.396 & 0.522 & 0.306 & 0.185 & 0.238 & 0.088 & 0.366 & 0.470 & 0.257 & 0.394  & 0.527  & 0.302  & 0.219  & 0.275  & 0.111 & 0.190  & 0.066  & 0.017 \\ \hline
		DCCM    & 0.496  &   0.623   & 0.408   & 0.285    & 0.327   & 0.173        &  0.376   & 0.482 &  0.262     &   0.608   & 0.710 & 0.555     &   0.321      &  0.383 &  0.182    &   0.224     &  0.108  & 0.038  \\ \hline
		IIC  & -  &    0.617  & -    & -     & 0.257     &  -   &  -   & 0.610 & - &  - & - &   -     &     -      &  - &  -    &   -     &  -   &  -   \\ \hline
		PICA  & 0.591   &   0.696    & 0.512  & 0.310  & 0.337   & 0.171   &  0.611  & 0.713 &  0.531 &  0.802    & 0.870 &   0.761   &   0.352  & 0.352 &  0.201   &  0.277   &  0.098    & 0.040   \\ \hline
	    DRC  & 
	    \underline{0.621}       &  \underline{0.727}  &   \underline{0.547}    & 
	    \underline{0.356}       &  \underline{0.367}  &   \underline{0.208}    &  
	    \underline{0.644}       &  \underline{0.747}  &   \underline{0.569}    &     
	    \underline{0.830}       &  \underline{0.884}  &   \underline{0.798}    &  
	    \underline{0.384}       &  \underline{0.389}  &   \underline{0.233}    &    
	    \underline{0.321}       &  \textbf{0.139}  &   \underline{0.056}   \\ \hline
	    
	    
	    \textbf{GCC} &
		$\bm{0.764}$      &   $\bm{0.856}$     &   $\bm{0.728}$   & 
		$\bm{0.472}$     &   $\bm{0.472}$     &   $\bm{0.305}$    & 
		$\bm{0.684}$     &   $\bm{0.788}$     &   $\bm{0.631}$   &     
		$\bm{0.842}$     &  $\bm{0.901}$      &   $\bm{0.822}$          &      
	    $\bm{0.490}$     &   $\bm{0.526}$     &   $\bm{0.362}$    &    
		$\bm{0.347}$      &   \underline{0.138}     &   $\bm{0.075}$ \\ \hline
	\end{tabular}
\end{table*}

\subsection{Experimental Settings}
\subsubsection{Datasets}
We conducted extensive experiments on six widely-adopted benchmark datasets. For a fair comparison, we adopted the same experimental setting as ~\cite{dac,pica}. The characteristics of these datasets are introduced in the following.

\begin{itemize}
\item \textbf{CIFAR-10/100:}~\cite{cifar10} The image size is $32\times32\times3$ and 10 classes and 20 super-classes are considered for the CIFAR-10/CIFAR-100 dataset in experiments. 50,000/10,000 training and testing images of each dataset are jointly utilized to clustering. 

\item \textbf{STL-10:}~\cite{stl10} The STL-10 is an image recognition dataset containing 500/800 training/test images for each of 10 classes with image size $96\times96\times3$ and additional 100,000 samples from several unknown classes for training stage. 

\item \textbf{ImageNet-10 and ImageNet-Dogs:}~\cite{dac} Two subsets of ImageNet~\cite{imagenet}: the former contains 10 randomly selected subjects and the latter contains 15 dog breeds. Their size is set to $96\times96\times3$.

\item \textbf{Tiny-ImageNet:}~\cite{tiny_imagenet} It is a very challenging tiny ImageNet dataset for clustering with 200 classes. There are 100,000/10,000 training/test images with dimension $64\times64\times3$ in each category. 
\end{itemize}

\subsubsection{Evaluation Metrics} 
Similar to~\cite{pica}, we adopted three standard metrics for evaluating the performance of clustering, including Accuracy~(ACC), Normalized Mutual Information~(NMI), and Adjusted Rand Index~(ARI).

\subsubsection{Compared Methods} 
We compared the proposed method with both traditional and deep learning based methods, including K-means, spectral clustering~(SC)~\cite{SpectralClustering}, agglomerative clustering~(AC)~\cite{gowda1978agglomerative}, the nonnegative matrix factorization~(NMF) based clustering~\cite{cai2009locality}, auto-encoder~(AE)~\cite{bengio2007greedy}, denoising auto-encoder~(DAE)~\cite{vincent2010stacked}, GAN~\cite{radford2015unsupervised}, deconvolutional networks~(DECNN)~\cite{zeiler2010deconvolutional}, variational auto-encoding~(VAE)~\cite{kingma2013auto}, deep embedding clustering~(DEC)~\cite{xie2016unsupervised}, jointly unsupervised learning~(JULE)~\cite{yang2016joint}, deep adaptive image clustering~(DAC)~\cite{dac}, invariant information clustering~\cite{iic}, deep comprehensive correlation Mining~(DCCM)~\cite{dccm}, partition confidence maximisation~(PICA)~\cite{pica}, and deep robust clustering~(DRC)~\cite{drc}.

\subsubsection{Implementation Details}
We utilized PyTorch~\cite{pytorch} to implement all experiments.
In our framework, we used ResNet-18~\cite{resnet} as the main network architecture and train networks on one Tesla P100 GPU. The SGD optimizer is adopt with $lr=0.4$, a weight decay $1e-4$ and momentum coefficient $0.9$. The learning rate decays by cosine scheduler with decay rate $0.1$. The batch size is set to 256 and the same data augmentation is adopted as~\cite{simCLR}\{color jitter, random grayscale, randomly resized crop\}.  The temperatures in RGC and AGC are set to $\tau=0.1$ and $\tau=1.0$, respectively. For hyper-parameters, we set $\alpha=0.5$, $\lambda=0.5$ and $\eta = 1.0$ for all datasets. 
For the construction of KNN graph, we set $K=5$
and utilized the efficient similarity search library 'Faiss'~\footnote{https://github.com/facebookresearch/faiss}. Even for 1 million samples with $256$ dimensional features on a CPU with $64$ cores and $2.5$GHz, it takes about $50$ seconds to construct a KNN graph .
Therefore, its time cost is neglectable and the KNN graph construction does not limit its application to large scale datasets.
For the ablation study, we adopted the same setting as SCAN~\cite{scan} to perform self-labeling processing. 

\subsection{Experimental Results and Analysis} 
In Table~\ref{tab:clustering_res}, we presented the clustering results of GCC and other related methods on these six challenging datasets. The results of other methods are directly copied from DRC~\cite{drc}. 
Based on the results, we can first see that deep learning based methods achieve much better results than traditional clustering methods due to the large parameter capacity. For instance, the accuracy of most deep learning based clustering methods on CIFAR-10 is much higher than 0.3, while the accuracy of these classic methods, including SC, AC, and NMF, is lower than 0.25.
Secondly, these contrastive learning based methods, such as PICA, DRC and GCC, are more suitable for the clustering task since they can learn more discriminative feature representation.
Most importantly, it is obvious that our GCC significantly surpasses other methods by a large margin on most benchmarks under three different evaluation metrics. 
Even compared with the recent state-of-the-art methods PICA and DRC, the improvement of GCC is also remarkable. Take the clustering accuracy for example, our results are 12.9\%, 10.5\%, 4.1\% higher than that of the second best method DRC on CIFAR-10, CIFAR-100 and STL-10, respectively. The above results can well demonstrate the effectiveness and robustness of our proposed method.

\subsection{Ablation Study}
According to the objective function in Eq.~\eqref{equ:obj}, there are three different losses in total.
In this section, we will demonstrate that RGC loss in Eq.~\eqref{equ:rgcloss}, AGC loss in Eq.~\eqref{equ:agcloss}, and cluster regularization loss in Eq.~\eqref{equ:crloss} are all very important to improve the performance. We will also evaluate the influence of a post-processing strategy used in SCAN~\cite{scan} and the superiority of graph contrastive for clustering-oriented representation learning over the basic contrastive learning method. 

\subsubsection{Effect of Graph Contrastive Loss}
We first investigated how RGC and AGC losses affect the clustering performance on CIFAR-10, CIFAR-100 and ImageNet-10. Results are shown in Table~\ref{tab:ablation_2}. Compared with the method without using graph contrastive, both RGC and AGC improve the clustering results on all three datasets, especially on CIFAR-10. All best results are achieved by GCC, which implies that both RGC and ARC terms are indispensable.

\begin{table}[!tp]
	\centering
	\caption{Effect of two graph contrastive losses, where $\checkmark$ means using graph information. Metric: ACC.}
	\label{tab:ablation_2}
	\begin{minipage}[t]{0.5\textwidth}
	    \centering
		\begin{tabular}{ccccc}
			\hline
            RGC &   AGC    &   CIFAR-10	&   CIFAR-100 	&   ImageNet-10      \\ \hline
             &       &   0.752       &   0.438       &   0.878 \\
            \checkmark &         &   0.809       &   0.463       &   0.884 \\
            &      \checkmark   &   0.825       &   0.462       &   0.893 \\
           \checkmark &     \checkmark    &   $\bm{0.856}$       &   $\bm{0.472}$       &   $\bm{0.901}$ \\
             \hline
		\end{tabular}\\[4pt] 
	\end{minipage}
\end{table}

\subsubsection{Effect of Cluster Regularization Loss}
Deep clustering methods can easily fall into a local optimal solution when most samples are assigned to the same cluster. We examined how the cluster regularization loss addresses this problem. As shown in Table~\ref{tab:ablation_3}, we can see that it significantly helps to improve the clustering performance. It is interesting to see the cluster regularization loss has little impact on ImageNet-10 since it is a relatively easy dataset where images from different classes are well separated.
\begin{table}[!tp]
	\centering
	\caption{Effect of cluster regularization loss. Metric: ACC.}
	\label{tab:ablation_3}
	\begin{minipage}[t]{0.5\textwidth}
	    \centering
		\begin{tabular}{cccc}
			\hline
            Method      &   CIFAR-10	&   CIFAR-100 	&   ImageNet-10      \\ \hline
            GCC  \textit{w/o} CR       &   0.680       &   0.348       &   0.828 \\ 
             GCC         &   $\bm{0.856}$       &   $\bm{0.472}$       &   $\bm{0.901}$ \\
             \hline
		\end{tabular}\\[4pt] 
	\end{minipage}
\end{table}

\subsubsection{Effect of Self-labeling Fine-tuning}
SCAN~\cite{scan} proposes a three-stage method for image clustering and achieved high performance. The clustering results benefit a lot by fine-tuning through self-labeling. For a fair comparison, we also performed self-labeling after GCC and the results are shown in~Table~\ref{tab:ablation_4}. We can see that GCC outperforms SCAN~\cite{scan} both before and after self-labeling on all three datasets reported in the paper of SCAN, which indicates that GCC learns more clustering-friendly representations and better clustering assignments.
\begin{table}[!tp]
	\centering
	\caption{Effect of self-labeling. * means that adopting self-label post-processing. Metric: ACC.}
	\label{tab:ablation_4}
	\begin{minipage}[t]{0.5\textwidth}
	    \centering
		\begin{tabular}{cccc}
			\hline
            Method    &   CIFAR-10	&   CIFAR-100 	&   STL-10      \\ \hline
            SCAN      &   0.818       &   0.422       &   0.755 \\
            GCC       &   $\bm{0.856}$       &   $\bm{0.472}$       &   $\bm{0.788}$   \\ \hline
            SCAN$^{*}$  &   0.883       &   0.507       &   0.809  \\
            GCC$^{*}$   &  $\bm{0.901}$ &  $\bm{ 0.523}$ &   $\bm{0.833}$ \\
             \hline
		\end{tabular}\\[4pt] 
	\end{minipage}
\end{table}
\begin{table}[!tp]
	\centering
	\caption{Comparison of features learned by GCC and simCLR. Metric: ACC.}
	\label{tab:ablation_5}
	\begin{minipage}[t]{0.5\textwidth}
	    \centering
		\begin{tabular}{cccc}
			\hline
            Method      &   CIFAR-10	&   CIFAR-100      \\ \hline
            simCLR + SC   &       0.660 &              0.292                      \\
            GCC + SC       &      \textbf{0.746} &                     \textbf{0.367}                 \\
            \hline
            simCLR + K-means         &   0.628       &   0.380 \\ 
            GCC + K-means         &   $\bm{0.754}$       &   $\bm{0.420}$  \\
            \hline
		\end{tabular}\\[4pt] 
	\end{minipage}
\end{table}
\begin{figure}[!tp]
\vspace{-1.5em}
\centering	
    \subfigure[Training Accuracy of Top-5 NN]{  
		\begin{minipage}[t]{0.22\textwidth}
			\centering    
			\includegraphics[width=4.3cm, height=3.8cm]{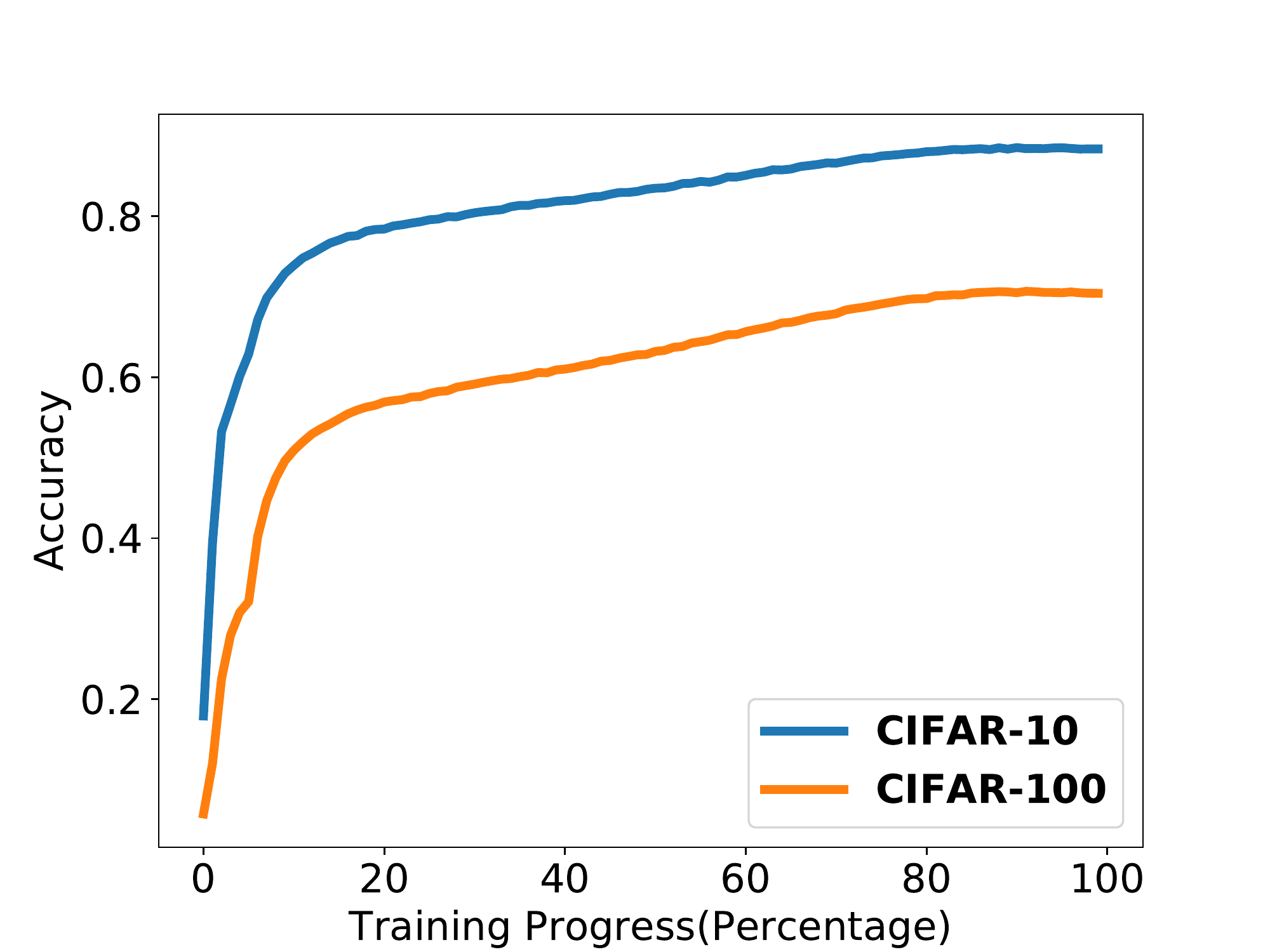}
		\end{minipage}
	}
	\subfigure[Top-K NN Accuracy]{  
		\begin{minipage}[t]{0.22\textwidth}
			\centering   
			\includegraphics[width=4.3cm, height=3.8cm]{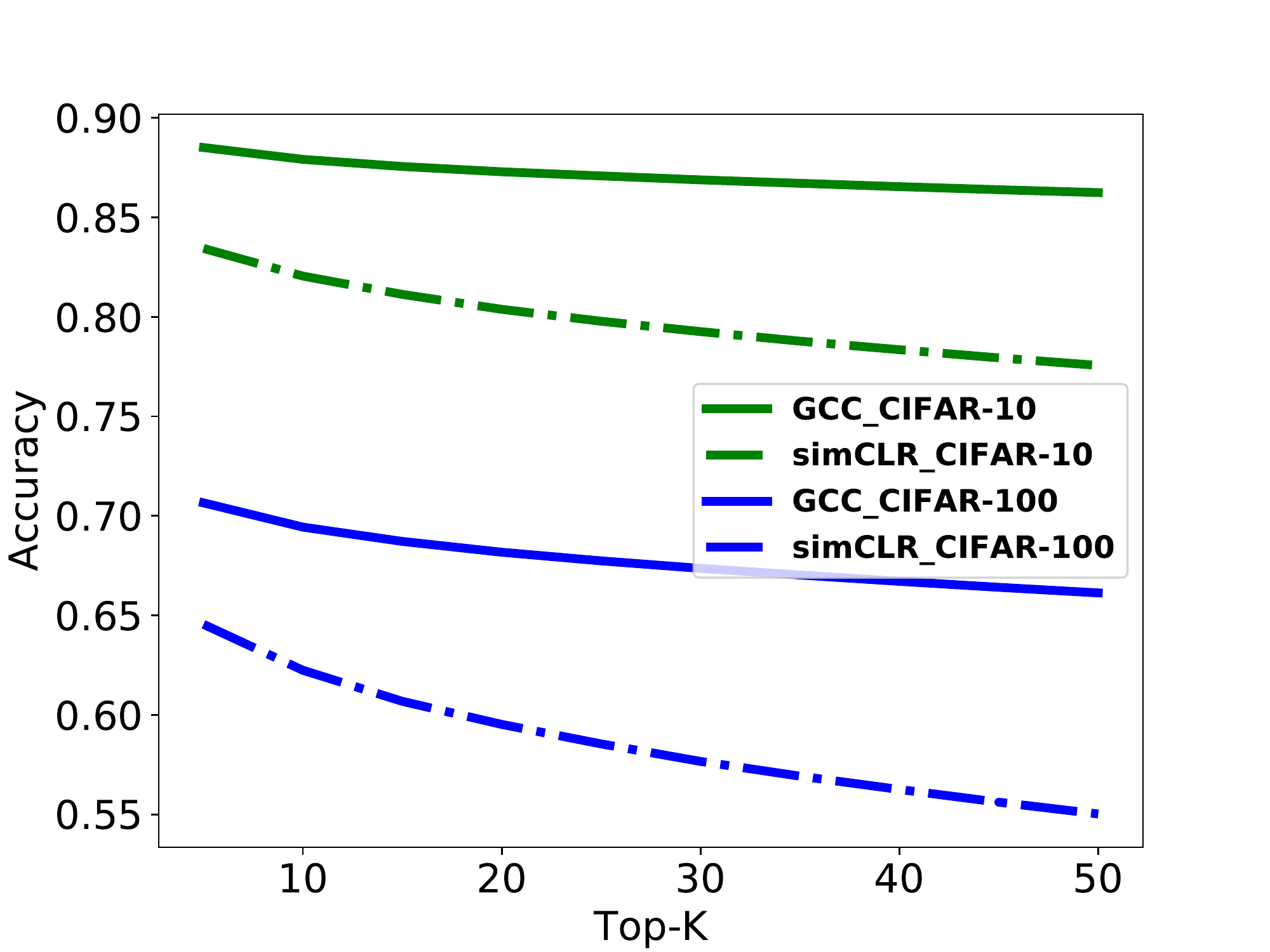}
		\end{minipage}  
	}
	\caption{Top-$K$ nearest neighbor accuracy of GCC and simCLR: (a) The evolution of top-5 NN accuracy for CIFAR-10 and CIFAR-100 during the training process of GCC. (b) The comparison of top-$K$ NN accuracy of CIFAR-10 and CIFAR-100 when varying $K$ from $1$ to $50$. 
	} 
	\label{fig:knn} 
\end{figure}

\begin{figure*}[!htbp]
	\centering
	\includegraphics[width=0.95\linewidth,height=4.5cm]{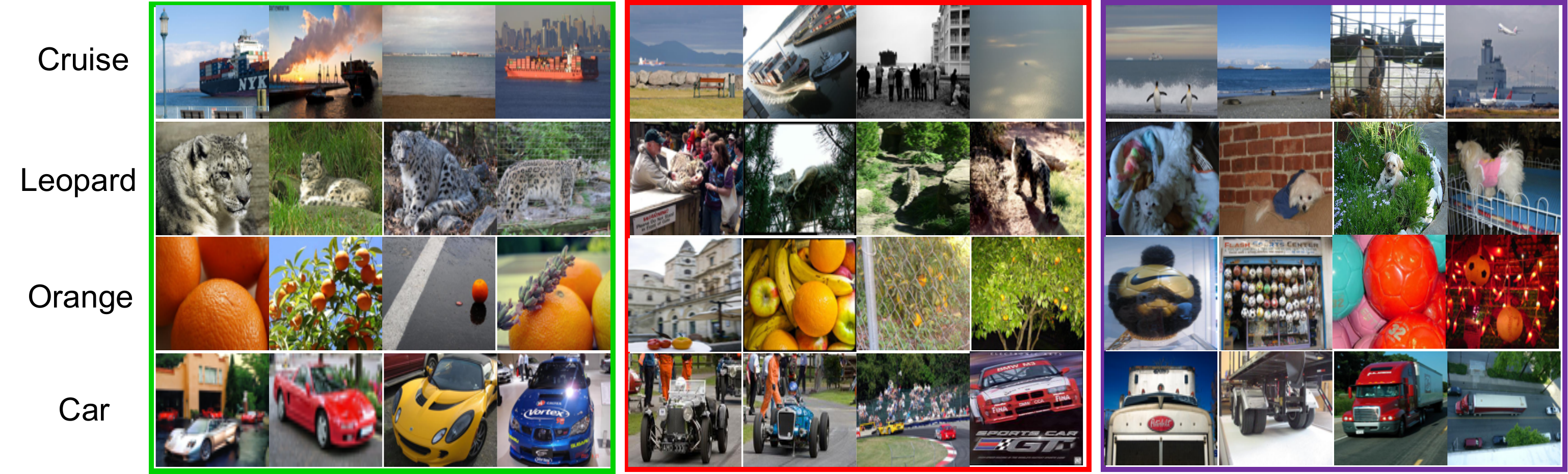}
	\caption{
	Case study on ImageNet-10. Successful cases~(left), false negative cases~(middle), and false positive failure cases~(right).
	} 
	\label{fig:vis} 
\end{figure*}

\subsubsection{Superiority of Graph Contrastive}
To demonstrate the superiority of Graph Contrastive on learned features, we performed two more quantitative analysis. First, we  directly adopted K-means and Spectral Clustering (SC)~\cite{SpectralClustering} to  cluster the learned features of basic contrastive learning (simCLR~\cite{simCLR}) and GCC on testing datasets (10,000 samples). For a fair comparison, here we only used RGC loss for GCC, and the implementation details are same to simCLR~\cite{simCLR}. As we can see from Table~\ref{tab:ablation_5}, the clustering performance of GCC is much better than simCLR, which verifies that the features learned by GCC are more conducive to clustering.

Furthermore, we calculated the accuracy of  top-$K$ nearest neighbor~(NN) obtained by GCC and simCLR, and the results are shown in Figure~\ref{fig:knn}. We can see that the top-$5$ NN accuracy of GCC becomes better and better during training from Figure~\ref{fig:knn}(a), which verifies the motivation of our graph contrastive learning. The comparison of GCC and simCLR are shown in Figure~\ref{fig:knn}(b), where the results of GCC are consistently better than simCLR when varying $K$ from $1$ to $50$.

\begin{table}[!tp]
	\centering
	\caption{Comparison of graph contrastive and ordinary contrastive learning with multiple positives. Metric: ACC.}
	\label{tab:ablation_6}
	\begin{minipage}[t]{0.5\textwidth}
	    \centering
		\begin{tabular}{cccc}
			\hline
            Method      &   CIFAR-10	&   CIFAR-100 	&   ImageNet-10      \\ \hline
            Multi-positive         &   0.807       &   0.426       &   0.872 \\ 
             GCC         &   $\bm{0.856}$       &   $\bm{0.472}$       &   $\bm{0.901}$ \\
             \hline
		\end{tabular}\\[4pt] 
	\end{minipage}
\end{table}

Several recent methods \cite{han2020self,khosla2020supervised} propose to extend basic contrastive learning by simply adding more positive samples. We replaced RGC with this contrastive loss to perform clustering analysis and the result is shown in Table~\ref{tab:ablation_6}. It is clear that GCC performs much better, which again demonstrates the advantages of our GC framework.

\subsection{Qualitative Study}
\subsubsection{Visualization of Representations}
\begin{figure}[!tp]
\vspace{-2em}
\centering
    \subfigure[Basic Contrastive Clustering]{  
		\begin{minipage}[t]{0.22\textwidth}
			\hspace*{-0.6cm}\includegraphics[width=5.2cm, height=4.0cm]{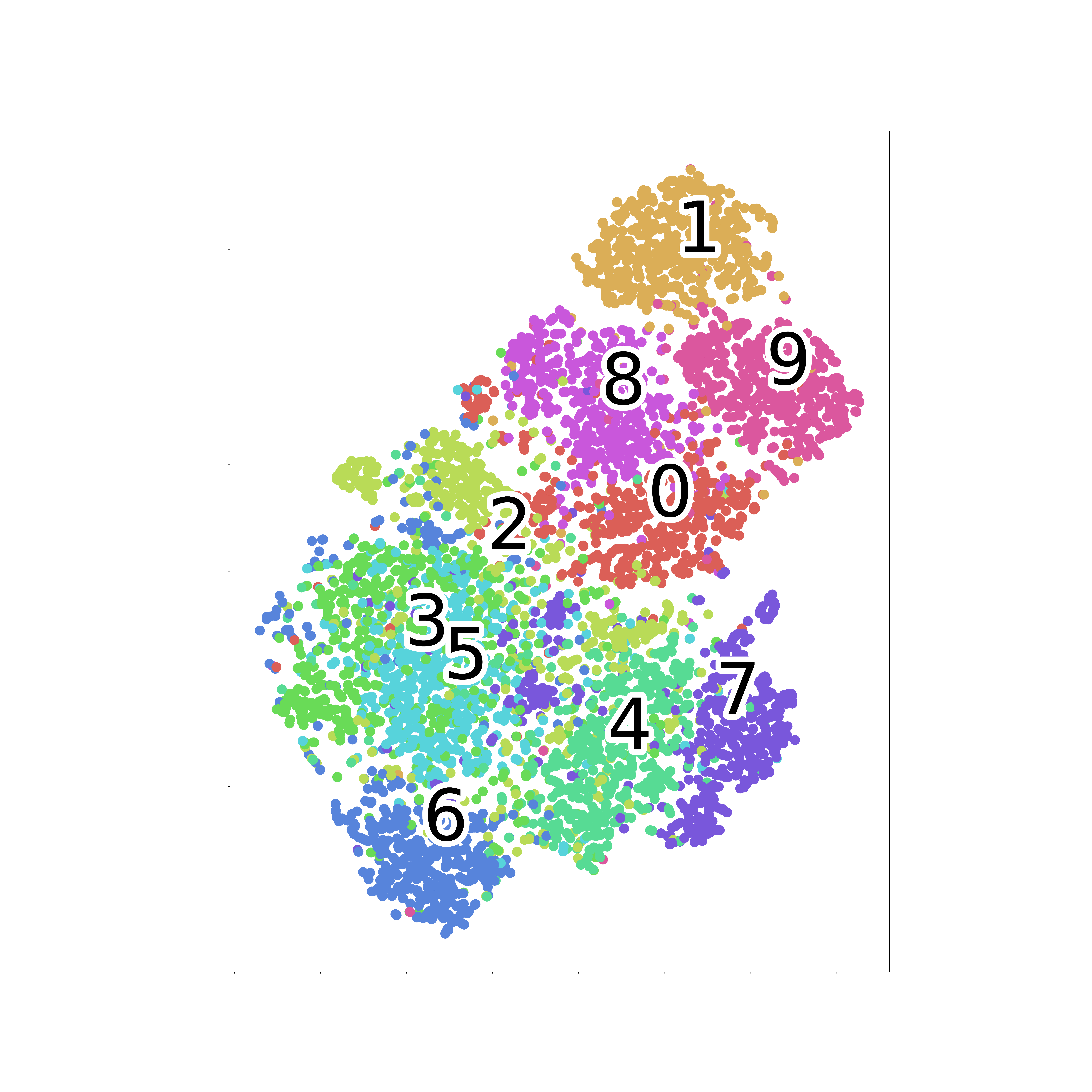}
		\end{minipage}
	}
	\subfigure[Graph Contrastive Clustering]{  
		\begin{minipage}[t]{0.22\textwidth}
			\hspace*{-0.4cm}\includegraphics[width=4.5cm,height=4.0cm]{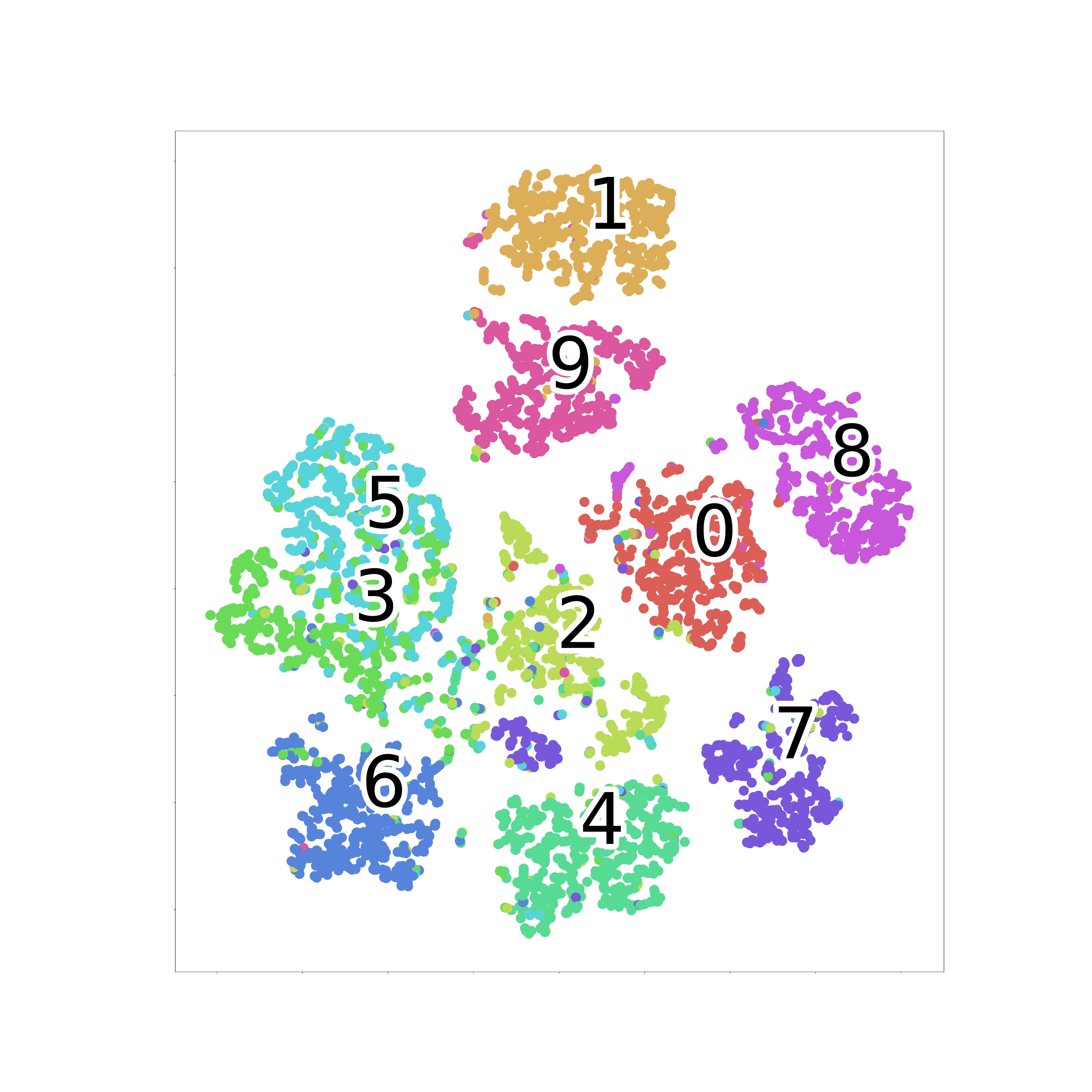}
		\end{minipage}  
	}
	\caption{t-SNE visualization for basic contrastive learning and our graph contrastive learning  on the CIFAR-10 dataset.
	} 
	\label{fig:tsne} 
	\vspace{-1em}
\end{figure}

To further illustrate that the features obtained by GCC are more suitable for clustering than simCLR, we visualized them on CIFAR-10 by t-SNE~\cite{maaten2008visualizing}. To be specific, we plotted the predictions of 6,000 randomly selected samples with the ground-truth classes color encoded by using t-SNE. As shown in Figure~\ref{fig:tsne}, samples in the same class are more compact and  samples of different classes are significantly better separated for GCC. For example, the samples of class 2 (in green-yellow) are divided into two parts in simCLR but gathered together in GCC.

\subsubsection{Case Study}

At last, we investigated both success and failure cases to get extra insights into our method. Specifically, we studied the following three cases of four classes from ImageNet-10: 
\textbf{(1)} Success cases, 
\textbf{(2)} False negative failure cases, 
\textbf{(3)} False positive cases. As shown in Figure \ref{fig:vis}, GCC can successfully group together images of the same class with different backgrounds and angles. Two different failure cases tell us that GCC mainly learns the shape of objects. Samples of different classes with a similar pattern may be grouped together and samples of the same class with different patterns may be separated into different classes. It is hard to look into the details at the absence of the ground-truth labels, which is still an unsolved problem for unsupervised learning.

\section{Conclusion}
To address the shortage of existing contrastive learning based clustering methods, we propose a novel graph contrastive learning framework, which is then applied to the clustering task and we come up with the Graph Constrastive Clustering~(GCC) method. Different from basic contrastive clustering that only maximizes the correlation between an image and its augmentation, we lift the instance-level feature consistency to the cluster-level consistency with the assumption that samples in one cluster and their augmentations should have similar representations. We perform extensive experiments on six widely-adopted deep clustering benchmarks that demonstrate GCC learns more clustering-friendly representations than basic contrastive learning and outperforms a wide range of state-of-the-art methods.

{\small
\bibliographystyle{ieee_fullname}
\bibliography{egbib}

\begin{thebibliography}{10}\itemsep=-1pt

\bibitem{bengio2007greedy}
Yoshua Bengio, Pascal Lamblin, Dan Popovici, and Hugo Larochelle.
\newblock Greedy layer-wise training of deep networks.
\newblock In {\em NeurIPS}, pages 153--160, 2007.

\bibitem{cai2009locality}
Deng Cai, Xiaofei He, Xuanhui Wang, Hujun Bao, and Jiawei Han.
\newblock Locality preserving nonnegative matrix factorization.
\newblock In {\em IJCAI}, volume~9, pages 1010--1015, 2009.

\bibitem{dac}
Jianlong Chang, Lingfeng Wang, Gaofeng Meng, Shiming Xiang, and Chunhong Pan.
\newblock Deep adaptive image clustering.
\newblock In {\em IEEE ICCV}, pages 5879--5887, 2017.

\bibitem{simCLR}
Ting Chen, Simon Kornblith, Mohammad Norouzi, and Geoffrey Hinton.
\newblock A simple framework for contrastive learning of visual
  representations.
\newblock {\em arXiv preprint arXiv:2002.05709}, 2020.

\bibitem{chen2020big}
Ting Chen, Simon Kornblith, Kevin Swersky, Mohammad Norouzi, and Geoffrey
  Hinton.
\newblock Big self-supervised models are strong semi-supervised learners.
\newblock {\em arXiv preprint arXiv:2006.10029}, 2020.

\bibitem{chen2020improved}
Xinlei Chen, Haoqi Fan, Ross Girshick, and Kaiming He.
\newblock Improved baselines with momentum contrastive learning.
\newblock {\em arXiv preprint arXiv:2003.04297}, 2020.

\bibitem{stl10}
Adam Coates, Andrew Ng, and Honglak Lee.
\newblock An analysis of single-layer networks in unsupervised feature
  learning.
\newblock In {\em AISTATS}, pages 215--223, 2011.

\bibitem{depict}
K.~G. {Dizaji}, A. {Herandi}, C. {Deng}, W. {Cai}, and H. {Huang}.
\newblock Deep clustering via joint convolutional autoencoder embedding and
  relative entropy minimization.
\newblock In {\em IEEE ICCV}, pages 5747--5756, 2017.

\bibitem{elhamifar2009sparse}
Ehsan Elhamifar and Ren{\'e} Vidal.
\newblock Sparse subspace clustering.
\newblock In {\em IEEE CVPR}, pages 2790--2797, 2009.

\bibitem{gowda1978agglomerative}
K~Chidananda Gowda and G Krishna.
\newblock Agglomerative clustering using the concept of mutual nearest
  neighbourhood.
\newblock {\em Pattern Recognition}, 10(2):105--112, 1978.

\bibitem{guo2017improved}
Xifeng Guo, Long Gao, Xinwang Liu, and Jianping Yin.
\newblock Improved deep embedded clustering with local structure preservation.
\newblock In {\em IJCAI}, pages 1753--1759, 2017.

\bibitem{haeusser2018associative}
Philip Haeusser, Johannes Plapp, Vladimir Golkov, Elie Aljalbout, and Daniel
  Cremers.
\newblock Associative deep clustering: Training a classification network with
  no labels.
\newblock In {\em German Conference on Pattern Recognition}, pages 18--32,
  2018.

\bibitem{han2020self}
Tengda Han, Weidi Xie, and Andrew Zisserman.
\newblock Self-supervised co-training for video representation learning.
\newblock {\em arXiv preprint arXiv:2010.09709}, 2020.

\bibitem{MoCo}
Kaiming He, Haoqi Fan, Yuxin Wu, Saining Xie, and Ross Girshick.
\newblock Momentum contrast for unsupervised visual representation learning.
\newblock In {\em IEEE CVPR}, pages 9729--9738, 2020.

\bibitem{resnet}
Kaiming He, Xiangyu Zhang, Shaoqing Ren, and Jian Sun.
\newblock Deep residual learning for image recognition.
\newblock In {\em IEEE CVPR}, pages 770--778, 2016.

\bibitem{pica}
Jiabo Huang, Shaogang Gong, and Xiatian Zhu.
\newblock Deep semantic clustering by partition confidence maximisation.
\newblock In {\em IEEE CVPR}, pages 8849--8858, 2020.

\bibitem{iic}
Xu Ji, Jo{\~a}o~F Henriques, and Andrea Vedaldi.
\newblock Invariant information clustering for unsupervised image
  classification and segmentation.
\newblock In {\em IEEE ICCV}, pages 9865--9874, 2019.

\bibitem{khosla2020supervised}
Prannay Khosla, Piotr Teterwak, Chen Wang, Aaron Sarna, Yonglong Tian, Phillip
  Isola, Aaron Maschinot, Ce Liu, and Dilip Krishnan.
\newblock Supervised contrastive learning.
\newblock {\em arXiv preprint arXiv:2004.11362}, 2020.

\bibitem{kingma2013auto}
Diederik~P Kingma and Max Welling.
\newblock Auto-encoding variational bayes.
\newblock {\em arXiv preprint arXiv:1312.6114}, 2013.

\bibitem{cifar10}
Alex Krizhevsky, Geoffrey Hinton, et~al.
\newblock Learning multiple layers of features from tiny images.
\newblock 2009.

\bibitem{imagenet}
Alex Krizhevsky, Ilya Sutskever, and Geoffrey~E Hinton.
\newblock Imagenet classification with deep convolutional neural networks.
\newblock In {\em NeurIPS}, pages 1097--1105, 2012.

\bibitem{tiny_imagenet}
Ya Le and Xuan Yang.
\newblock Tiny imagenet visual recognition challenge.
\newblock {\em CS 231N}, 7, 2015.

\bibitem{cc}
Yunfan Li, Peng Hu, Zitao Liu, Dezhong Peng, Joey~Tianyi Zhou, and Xi Peng.
\newblock Contrastive clustering.
\newblock {\em arXiv preprint arXiv:2009.09687}, 2020.

\bibitem{liu2013robust}
Guangcan Liu, Zhouchen Lin, Shuicheng Yan, Ju Sun, Yong Yu, and Yi Ma.
\newblock Robust recovery of subspace structures by low-rank representation.
\newblock {\em IEEE TPAMI}, 35(1):171--184, 2013.

\bibitem{maaten2008visualizing}
Laurens van~der Maaten and Geoffrey Hinton.
\newblock Visualizing data using t-sne.
\newblock {\em Journal of Machine Learning Research}, 9(Nov):2579--2605, 2008.

\bibitem{ng2002spectral}
Andrew~Y Ng, Michael~I Jordan, and Yair Weiss.
\newblock On spectral clustering: Analysis and an algorithm.
\newblock In {\em NeurIPS}, pages 849--856, 2002.

\bibitem{pytorch}
Adam Paszke, Sam Gross, Soumith Chintala, Gregory Chanan, Edward Yang, Zachary
  DeVito, Zeming Lin, Alban Desmaison, Luca Antiga, and Adam Lerer.
\newblock Automatic differentiation in pytorch.
\newblock 2017.

\bibitem{peng2016deep}
Xi Peng, Shijie Xiao, Jiashi Feng, Wei-Yun Yau, and Yi Zhang.
\newblock Deep subspace clustering with sparsity prior.
\newblock In {\em IJCAI}, pages 1925--1931, 2016.

\bibitem{radford2015unsupervised}
Alec Radford, Luke Metz, and Soumith Chintala.
\newblock Unsupervised representation learning with deep convolutional
  generative adversarial networks.
\newblock {\em arXiv preprint arXiv:1511.06434}, 2015.

\bibitem{SpectralClustering}
Jianbo Shi and Jitendra Malik.
\newblock Normalized cuts and image segmentation.
\newblock {\em IEEE Transactions on Pattern Analysis and Machine Intelligence},
  22(8):888--905, 2000.

\bibitem{tian2019contrastive}
Yonglong Tian, Dilip Krishnan, and Phillip Isola.
\newblock Contrastive multiview coding.
\newblock {\em arXiv preprint arXiv:1906.05849}, 2019.

\bibitem{tian2020ICLR}
Yonglong Tian, Dilip Krishnan, and Phillip Isola.
\newblock Contrastive representation distillation.
\newblock In {\em ICLR}, 2020.

\bibitem{scan}
Wouter Van~Gansbeke, Simon Vandenhende, Stamatios Georgoulis, Marc Proesmans,
  and Luc Van~Gool.
\newblock Scan: Learning to classify images without labels.
\newblock In {\em ECCV}, 2020.

\bibitem{vincent2010stacked}
Pascal Vincent, Hugo Larochelle, Isabelle Lajoie, Yoshua Bengio, Pierre-Antoine
  Manzagol, and L{\'e}on Bottou.
\newblock Stacked denoising autoencoders: Learning useful representations in a
  deep network with a local denoising criterion.
\newblock {\em Journal of Machine Learning Research}, 11(12), 2010.

\bibitem{dccm}
Jianlong Wu, Keyu Long, Fei Wang, Chen Qian, Cheng Li, Zhouchen Lin, and
  Hongbin Zha.
\newblock Deep comprehensive correlation mining for image clustering.
\newblock In {\em IEEE ICCV}, pages 8150--8159, 2019.

\bibitem{wu2018unsupervised}
Zhirong Wu, Yuanjun Xiong, Stella~X Yu, and Dahua Lin.
\newblock Unsupervised feature learning via non-parametric instance
  discrimination.
\newblock In {\em IEEE CVPR}, pages 3733--3742, 2018.

\bibitem{xie2016unsupervised}
Junyuan Xie, Ross Girshick, and Ali Farhadi.
\newblock Unsupervised deep embedding for clustering analysis.
\newblock In {\em ICML}, pages 478--487, 2016.

\bibitem{yang2017towards}
Bo Yang, Xiao Fu, Nicholas~D Sidiropoulos, and Mingyi Hong.
\newblock Towards k-means-friendly spaces: Simultaneous deep learning and
  clustering.
\newblock In {\em ICML}, pages 3861--3870, 2017.

\bibitem{yang2016joint}
Jianwei Yang, Devi Parikh, and Dhruv Batra.
\newblock Joint unsupervised learning of deep representations and image
  clusters.
\newblock In {\em IEEE CVPR}, pages 5147--5156, 2016.

\bibitem{zeiler2010deconvolutional}
Matthew~D Zeiler, Dilip Krishnan, Graham~W Taylor, and Rob Fergus.
\newblock Deconvolutional networks.
\newblock In {\em IEEE CVPR}, pages 2528--2535, 2010.

\bibitem{zelnik2005self}
Lihi Zelnik-Manor and Pietro Perona.
\newblock Self-tuning spectral clustering.
\newblock In {\em NeurIPS}, pages 1601--1608, 2005.

\bibitem{drc}
Huasong Zhong, Chong Chen, Zhongming Jin, and Xian-Sheng Hua.
\newblock Deep robust clustering by contrastive learning.
\newblock {\em arXiv preprint arXiv:2008.03030}, 2020.

\bibitem{zhuang2019local}
Chengxu Zhuang, Alex~Lin Zhai, and Daniel Yamins.
\newblock Local aggregation for unsupervised learning of visual embeddings.
\newblock In {\em IEEE ICCV}, pages 6002--6012, 2019.

\end{thebibliography}
}

\end{document}